\documentclass[runningheads]{llncs}

\usepackage{graphicx}

\usepackage{comment}
\usepackage{amsmath,amssymb} %
\usepackage{color}
\usepackage{booktabs} %
\usepackage[dvipsnames]{xcolor}
\usepackage{multirow}
\usepackage{array}
\usepackage{tabularx}
\usepackage{mathrsfs}
\usepackage{bm}
\DeclareUnicodeCharacter{2212}{-}
\usepackage[ruled,vlined]{algorithm2e}
\usepackage{caption}
\usepackage{colortbl}
\usepackage{hyperref}
\hypersetup{colorlinks,linkcolor={blue},citecolor={green},urlcolor={red}} 
\usepackage[accsupp]{axessibility}  %
\usepackage[labelfont=bf]{caption}
\usepackage[nice]{nicefrac}

\definecolor{ForestGreen}{RGB}{34,139,34}
\definecolor{LightCyan}{rgb}{0.88,1,1}

\newcommand{\myparagraph}[1]{\vspace{1pt}\noindent{\bf{#1}}}

\definecolor{Gray1}{gray}{0.85}
\definecolor{Gray2}{gray}{0.9}
\definecolor{Gray3}{gray}{0.99}
\definecolor{LightCyan}{rgb}{0.88,1,1}
\newcolumntype{a}{>{\columncolor{Gray1}}c}
\newcolumntype{b}{>{\columncolor{Gray2}}c}
\newcolumntype{d}{>{\columncolor{LightCyan}}c}
\newcolumntype{e}{>{\columncolor{Gray3}}c}

\usepackage[symbol]{footmisc}

\begin{document}
\pagestyle{headings}
\mainmatter
\def\ECCVSubNumber{1824}  %

\title{BayesCap: Bayesian Identity Cap for Calibrated Uncertainty in Frozen Neural Networks} %

\titlerunning{\texttt{BayesCap}: Bayesian Identity Cap for Fast Calibrated Uncertainty}
\author{Uddeshya Upadhyay$^*$\inst{1} \and
Shyamgopal Karthik$^*$\inst{1} \and
Yanbei Chen\inst{1} \and\\
Massimiliano Mancini\inst{1} \and
Zeynep Akata\inst{1,2}}
\authorrunning{U. Upadhyay et al.}
\institute{$^1$University of T\"{u}bingen $^2$Max Planck Institute for Intelligent Systems}

\maketitle

\vspace{-20pt}
\begin{abstract}
High-quality calibrated uncertainty estimates are crucial for numerous real-world applications, especially for deep learning-based deployed ML systems. While Bayesian deep learning techniques allow uncertainty estimation, training them with large-scale datasets is an expensive process that does not always yield models competitive with non-Bayesian counterparts. Moreover, many of the high-performing deep learning models that are already trained and deployed are non-Bayesian in nature and do not provide uncertainty estimates. 
To address these issues, we propose \texttt{BayesCap} that learns a Bayesian identity mapping for the frozen model, allowing uncertainty estimation. \texttt{BayesCap} is a memory-efficient method that can be trained on a small fraction of the original dataset, enhancing pretrained non-Bayesian computer vision models by providing calibrated uncertainty estimates for the predictions without (i) hampering the performance of the model and (ii) the need for expensive retraining the model from scratch. The proposed method is agnostic to various architectures and tasks. We show the efficacy of our method on a wide variety of tasks with a diverse set of architectures, including image super-resolution, deblurring, inpainting, and crucial application such as medical image translation. Moreover, 
we apply the derived uncertainty estimates to detect out-of-distribution samples in critical scenarios like depth estimation in autonomous driving. Code is available at \url{https://github.com/ExplainableML/BayesCap}.
\end{abstract}

\vspace{-15pt}
\footnotetext[1]{Equal contribution}

\section{Introduction}
\label{sec:intro}
\begin{figure}[!t]
    \centering
    \includegraphics[width=\textwidth]{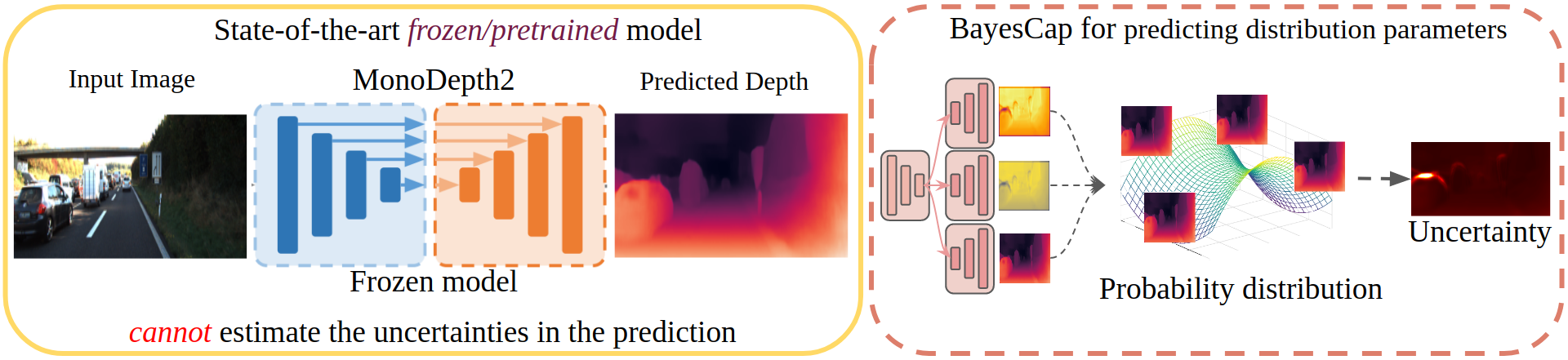}
    \vspace{-20pt}
    \caption{
        Computer vision models for image enhancements and translations deterministically map input to output without producing the uncertainty in the latter (example on the right shows depth estimation using MonoDepth2~\cite{godard2019digging}). \texttt{BayesCap} approximates the underlying distribution and adds uncertainty to the predictions of pretrained models efficiently, details in Section~\ref{sec:bayescap}.
    }
    \vspace{-18pt}
    \label{fig:motivation}
\end{figure}

\vspace{-5pt}
Image enhancement and translation tasks like super-resolution~\cite{Ledig_2017_CVPR}, deblurring \cite{kupyn2018deblurgan,kupyn2019deblurgan}, inpainting~\cite{yu2018generative}, colorization~\cite{zhang2016colorful,iizuka2016let}, denoising~\cite{tian2020deep,plotz2017benchmarking}, medical image synthesis~\cite{zhu2020cross,armanious2020medgan,cohen2018distribution,upadhyay2019robust,upadhyay2019mixed,upadhyay2021uncertaintyICCV,sudarshan2021towards}, monocular depth estimation in autonomous driving~\cite{fu2018deep,godard2019digging}, etc., 
have been effectively tackled using deep learning methods generating high-fidelity outputs. 
But, the respective state-of-the-art models usually learn a \textit{deterministic} one-to-one mapping between the input and the output, %
without modeling the uncertainty in the prediction.
For instance, a depth estimation model predicts a depth map from the input RGB image (Figure~\ref{fig:motivation}-(Left)), without providing uncertainty.
In contrast, learning a probabilistic mapping between the input and the output yields the underlying distribution and provides uncertainty estimates for the predictions. This is a vital feature in safety-critical applications such as autonomous driving and medical imaging.
For instance, well-calibrated uncertainty estimates can be used to trigger human/expert intervention in highly uncertain predictions, consequently preventing fatal automated decision making~\cite{coglianese2016regulating,schwarting2018planning,van2021artificial}.
The conventional approach for obtaining uncertainty estimates is to train Bayesian models \textit{from scratch}. However, Bayesian deep learning techniques are difficult to train and are not scalable to large volumes of high-dimensional data~\cite{daxberger2021laplace}. Moreover, they %
cannot be easily integrated with
sophisticated deterministic architectures and training schemes tailored for specific tasks,  %
vital to achieving
state-of-the-art in vision applications~\cite{dosovitskiy2015flownet,godard2017unsupervised}.

To address the above challenges, we enhance the predictions of pretrained state-of-the-art non-Bayesian deterministic 
deep models with uncertainty estimation %
while preserving their strong model performances.
There is limited literature tackling the similar problem~\cite{wang2019aleatoric,ayhan2018test,daxberger2021laplace} but these methods do not yield well-calibrated uncertainty estimates or do not scale to high-dimensional cases such as image synthesis, translation, and enhancement. 

In this work, we propose \texttt{BayesCap}, shown in Figure \ref{fig:motivation}-(Right), an architecture agnostic, plug-and-play method to generate uncertainty estimates for pretrained models. The key idea is to train a Bayesian autoencoder over the output images of the pretrained network, approximating the underlying distribution for the output images. Due to its Bayesian design, in addition to reconstructing the input, \texttt{BayesCap} also estimates the parameters of the underlying distribution, allowing us to compute the uncertainties. \texttt{BayesCap} is highly data-efficient and can be trained on a small fraction of the original dataset. 
For instance, \texttt{BayesCap} is 3-5$\times$ faster to train as compared to a Bayesian model from scratch, while still achieving uncertainty estimates that are better calibrated than the baselines. 

To summarize, we make the following contributions. (1)~We propose \texttt{BayesCap}, a simple method for generating post-hoc uncertainty estimates, by learning a Bayesian identity mapping, over the outputs of image synthesis/translation tasks with deterministic pretrained models. (2)~\texttt{BayesCap} leads to calibrated uncertainties while retaining the performance of the underlying state-of-the-art pretrained network on a variety of tasks including super-resolution, deblurring, inpainting, and medical imaging. (3)~We also show that quantifying uncertainty using \texttt{BayesCap} can help in downstream tasks such as Out-of-Distribution (OOD) detection in critical applications like autonomous driving. 
\section{Related Works}
\label{sec:related}
\vspace{-5pt}

\textbf{Image Enhancement and Translations.} 
Advances in computer vision led to tackle challenging problems such as super-resolution~\cite{dong2015image,Ledig_2017_CVPR}, denoising~\cite{tian2020deep,plotz2017benchmarking}, deblurring \cite{nah2017deep,kupyn2018deblurgan,kupyn2019deblurgan}, inpainting~\cite{pathak2016context,yu2018generative}, depth estimation~\cite{fu2018deep,godard2019digging} among others. 
Such problems are tackled using a diverse set of architectures and learning schemes. For instance,
the popular method for super-resolution involves training a conditional \textit{generative adversarial networks} (GANs), where the generator is conditioned with a low-resolution image and employs a pretrained VGG network~\cite{vgg} to enforce the content loss in the feature space along with the adversarial term from the discriminator~\cite{Ledig_2017_CVPR}. 
Differently, for the inpainting task, \cite{yu2018generative} uses a conditional GAN with contextual attention and trains the network using spatially discounted reconstruction loss. 
In the case of monocular depth estimation, recent works exploit the left-right consistency as a cue to train the model in an unsupervised fashion \cite{godard2017unsupervised}. 
While these methods are highly diverse in their architectures, training schemes, supervisory signals, etc., they typically focus on providing a deterministic one-to-one mapping which may not be ideal in many critical scenarios such as autonomous driving~\cite{mcallister2017concrete} and medical imaging~\cite{upadhyay2021uncertainty,upadhyay2021uncertaintyICCV,upadhyay2021robustness}.
Our \texttt{BayesCap} preserves the high-fidelity outputs provided by such deterministic pretrained models while approximating the underlying distribution of the output of such models, allowing uncertainty estimation.

\textbf{Uncertainty Estimation.} 
Bayesian deep learning models are capable of estimating the uncertainties in their prediction %
~\cite{lakshminarayanan2016simple,kendall2017uncertainties}. 
Uncertainties can be broadly divided into two categories; (1)~Epistemic uncertainty which is the uncertainty due to the model parameters~\cite{graves2011practical,blundell2015weight,daxberger2021laplace,welling2011bayesian,chen2014stochastic,lakshminarayanan2016simple,gal2016dropout}. (2)~Aleatoric uncertainty which is the underlying uncertainty in the measurement itself, often estimated by approximating the per-pixel residuals between the predictions and the ground-truth using a \textit{heteroscedastic}  distribution whose parameters are predicted as the output of the network which is trained \textit{from scratch} to maximize the likelihood of the system~\cite{kendall2017uncertainties,litjens2017survey,bae2021estimating,wang2019aleatoric,wang2021bayesian,laves2020well}. 
While epistemic uncertainty is important in low-data regimes as parameter estimation becomes noisy, however, this is often not the case in computer vision settings with large scale datasets where aleatoric uncertainty is the critical quantity~\cite{kendall2017uncertainties}.
However, it is expensive to train these models and they often 
perform worse than their deterministic counterparts~\cite{osawa2019practical,riquelme2018deep,daxberger2021laplace}.
Unlike these works, \texttt{BayesCap} is a fast and efficient method to estimate uncertainty over the %
predictions of a pretrained deterministic model.

\textbf{Post-hoc Uncertainty Estimation.}
 While this has not been widely explored, some recent works~\cite{daxberger2021laplace,eschenhagen2021mixtures} have tried to use the Laplace approximation for this purpose. However, these methods computes the Hessian which is not feasible for high-dimensional modern problems in computer vision~\cite{zhang2016colorful,pathak2016context,fu2018deep,kupyn2018deblurgan,plotz2017benchmarking}. Another line of work to tackle this problem is test-time data augmentation~\cite{wang2019aleatoric,ayhan2018test} that perturbs the inputs to obtain multiple outputs leading to uncertainties. However, these estimates are often poorly calibrated~\cite{gawlikowski2021survey}.
It is of paramount importance that the uncertainty estimates are well calibrated~\cite{kuleshov2018accurate,guo2017calibration,laves2020calibration,laves2020well,phan2018calibrating,zhang2020mix}. In many high-dimensional computer vision problems the per-pixel output is often a continuous value~\cite{pathak2016context,kupyn2018deblurgan,Ledig_2017_CVPR,zhang2016colorful}, i.e., the problem is regression in nature.
Recent works focused on \textit{Uncertainty Calibration Error} that generalizes to high dimensional regression~\cite{kuleshov2018accurate,levi2019evaluating,laves2020calibration,laves2020well}. 
Unlike prior works~\cite{ayhan2018test,daxberger2021laplace,wang2019aleatoric,eschenhagen2021mixtures}, \texttt{BayesCap} scales to high-dimensional tasks, %
providing well-calibrated uncertainties.%

\section{Methodology: \texttt{BayesCap} - Bayesian Identity Cap}
\vspace{-5pt}
We first describe the problem formulation in Section \ref{sec:prelim}, and preliminaries on uncertainty estimation in Section \ref{sec:pre_bayescap}. 
In Section~\ref{sec:bayescap}, we describe construction of \texttt{BayesCap} that models a probabilistic identity function capable of estimating the high-dimensional complex distribution %
from the frozen deterministic model, estimating calibrated uncertainty for the predictions. 

\subsection{Problem formulation}%
\label{sec:prelim}

Let $\mathcal{D} = \{(\mathbf{x}_i, \mathbf{y}_i)\}_{i=1}^{N}$ be the training set with pairs from domain $\mathbf{X}$ and $\mathbf{Y}$ (i.e., $\mathbf{x}_i \in \mathbf{X}, \mathbf{y}_i \in \mathbf{Y}, \forall i$), where $\mathbf{X}, \mathbf{Y}$ lies in $\mathbb{R}^m$ and $\mathbb{R}^n$, respectively. 
While our proposed solution is valid for data of arbitrary dimension, we present the formulation for images with applications for image enhancement and translation tasks, such as super-resolution, inpainting, etc. 
Therefore, ($\mathbf{x}_i, \mathbf{y}_i$) represents a pair of images, where $\mathbf{x}_i$ refers to the input and $\mathbf{y}_i$ denotes the transformed/enhanced output. 
For instance, in super-resolution $\mathbf{x}_i$ is a low-resolution image and $\mathbf{y}_i$ its high-resolution version.
Let $\mathbf{\Psi}(\cdot; \theta): \mathbb{R}^m \rightarrow \mathbb{R}^n$ represent a Deep Neural Network parametrized by $\theta$ that maps images from the set $\mathbf{X}$ to the set $\mathbf{Y}$, e.g. from corrupted to the non-corrupted/enhanced output images. 

We consider a real-world scenario, where $\mathbf{\Psi}(\cdot; \theta)$ has already been trained using the dataset $\mathcal{D}$ and it is in a \textit{frozen state} with parameters set to the learned optimal parameters $\theta^{*}$. In this state, given an input %
$\mathbf{x}$, the model returns %
a point estimate of the output, i.e., $\hat{\mathbf{y}} = \mathbf{\Psi}(\mathbf{x}; \theta^{*})$.
However, %
point estimates %
do not %
capture the distributions of the output ($\mathcal{P}_{\mathbf{Y}|\mathbf{X}}$) and thus %
the uncertainty in the prediction that is crucial in many real-world  applications~\cite{kendall2017uncertainties}.
Therefore, we propose to estimate $\mathcal{P}_{\mathbf{Y}|\mathbf{X}}$ for the pretrained model in a fast and cheap manner, quantifying the uncertainties of the output without re-training the model itself. 

\subsection{Preliminaries: Uncertainty Estimation} %
\label{sec:pre_bayescap}
To understand the functioning of our \texttt{BayesCap} that produces uncertainty estimates for the \textit{frozen or pretrained} neural networks, we first %
consider a model %
trained from scratch to address the target task and estimate uncertainty. Let us denote this model by $\mathbf{\Psi}_s(\cdot; \zeta): \mathbb{R}^m \rightarrow \mathbb{R}^n$, with a set of trainable parameters given by $\zeta$. To capture the \textit{irreducible} (i.e., aleatoric) uncertainty in the output distribution $\mathcal{P}_{Y|X}$, the model must estimate the parameters of the distribution. These are then used to maximize the likelihood function. 
That is,
for an input $\mathbf{x}_i$, the model produces a set of parameters representing the output given by, $\{\hat{\mathbf{y}}_i, \hat{\nu}_i \dots \hat{\rho}_i \} := \mathbf{\Psi}_s(\mathbf{x}_i; \zeta)$, that characterizes the distribution 
$\mathcal{P}_{Y|X}(\mathbf{y}; \{\hat{\mathbf{y}}_i, \hat{\nu}_i \dots \hat{\rho}_i \})$, such that 
$\mathbf{y}_i \sim \mathcal{P}_{Y|X}(\mathbf{y}; \{\hat{\mathbf{y}}_i, \hat{\nu}_i \dots \hat{\rho}_i \})$.
The likelihood $\mathscr{L}(\zeta; \mathcal{D}) := \prod_{i=1}^{N} \mathcal{P}_{Y|X}(\mathbf{y}_i; \{\hat{\mathbf{y}}_i, \hat{\nu}_i \dots \hat{\rho}_i \})$ is then maximized in order to estimate the optimal parameters of the network. Moreover, the distribution $\mathcal{P}_{Y|X}$ is often chosen such that uncertainty can be estimated using a closed form solution $\mathscr{F}$ depending on the estimated parameters %
of the neural network, i.e.,
\begin{gather}
\{\hat{\mathbf{y}}_i, \hat{\nu}_i \dots \hat{\rho}_i \} := \mathbf{\Psi}_s(\mathbf{x}_i; \zeta) \\
\zeta^* := \underset{\zeta}{\text{argmax }} \mathscr{L}(\zeta; \mathcal{D}) = \underset{\zeta}{\text{argmax}} \prod_{i=1}^{N} \mathcal{P}_{Y|X}(\mathbf{y}_i; \{\hat{\mathbf{y}}_i, \hat{\nu}_i \dots \hat{\rho}_i \}) \\
\text{Uncertainty}(\hat{\mathbf{y}}_i) = \mathscr{F}(\hat{\nu}_i \dots \hat{\rho}_i)
\end{gather}
It is common to use a \textit{heteroscedastic} Gaussian distribution for $\mathcal{P}_{Y|X}$~\cite{kendall2017uncertainties,wang2019aleatoric}, in which case $\mathbf{\Psi}_s(\cdot; \zeta)$ is designed to predict the \textit{mean} and \textit{variance} of the Gaussian distribution, i.e., 
$\{\hat{\mathbf{y}}_i, \hat{\sigma}_i^2 \} := \mathbf{\Psi}_s(\mathbf{x}_i; \zeta)$,
and the predicted \textit{variance} itself can be treated as uncertainty in the prediction. The optimization problem becomes,
\begin{gather}
\zeta^* %
= \underset{\zeta}{\text{argmax}} \prod_{i=1}^{N} \frac{1}{\sqrt{2 \pi \hat{\sigma}_i^2}} 
e^{-\frac{|\hat{\mathbf{y}}_i - \mathbf{y}_i|^2}{2\hat{\sigma}_i^2}} 
= 
\underset{\zeta}{\text{argmin}} \sum_{i=1}^{N} \frac{|\hat{\mathbf{y}}_i - \mathbf{y}_i|^2}{2\hat{\sigma}_i^2} + \frac{\log(\hat{\sigma}_i^2)}{2} \label{eq:scratch_gauss}\\
\text{Uncertainty}(\hat{\mathbf{y}}_i) = \hat{\sigma}_i^2.
\end{gather}

The above equation models the per-pixel residual (between the prediction and the ground-truth) as a Gaussian distribution. However, this may not always be fit, especially in the presence of outliers and artefacts, where the residuals often follow heavy-tailed distributions.
Recent works such as~\cite{upadhyay2021robustness,upadhyay2021uncertaintyICCV} have shown that heavy-tailed distributions can be modeled as a heteroscedastic generalized Gaussian distribution, in which case $\mathbf{\Psi}_s(\cdot; \zeta)$ is designed to predict the \textit{mean} ($\hat{\mathbf{y}}_i$), \textit{scale} ($\hat{\mathbf{\alpha}}_i$), and \textit{shape} ($\hat{\mathbf{\beta}}_i$) as trainable parameters, i.e., 
$\{\hat{\mathbf{y}}_i, \hat{\alpha}_i, \hat{\beta}_i \} := \mathbf{\Psi}_s(\mathbf{x}_i; \zeta)$, 
\begin{gather}
\zeta^*
:= \underset{\zeta}{\text{argmax }} \mathscr{L}(\zeta)
= \underset{\zeta}{\text{argmax}} \prod_{i=1}^{N} \frac{\hat{\beta}_i}{2 \hat{\alpha}_i \Gamma(\frac{1}{\hat{\beta}_i})} 
e^{-(|\hat{\mathbf{y}}_i - \mathbf{y}_i|/\hat{\alpha}_i)^{\hat{\beta}_i}} 
= \underset{\zeta}{\text{argmin}} -\log\mathscr{L}(\zeta)
\nonumber \\ 
= 
\underset{\zeta}{\text{argmin}} \sum_{i=1}^{N} 
\left(
\frac{|\hat{\mathbf{y}}_{i}-\mathbf{y}_{i}|}{\hat{\alpha}_{i}} \right)^{\hat{\beta}_{i}} - 
    \log\frac{\hat{\beta}_{i}}{\hat{\alpha}_{i}} + \log\Gamma(\frac{1}{\hat{\beta}_{i}})
\\
\text{Uncertainty}(\hat{\mathbf{y}}_i) = \frac{\hat{\alpha}_i^2\Gamma(\frac{3}{\hat{\beta}}_i)}{\Gamma(\frac{1}{\hat{\beta}}_i)}.
\label{eq:sratch_ggd}
\end{gather}
Here $\Gamma(z) = \int_{0}^{\infty}x^{z-1}e^{-x} dx \text{, } \forall z>0$, represents the Gamma function~\cite{artin2015gamma}.
\begin{figure}[!t]
    \centering
    \includegraphics[width=\textwidth]{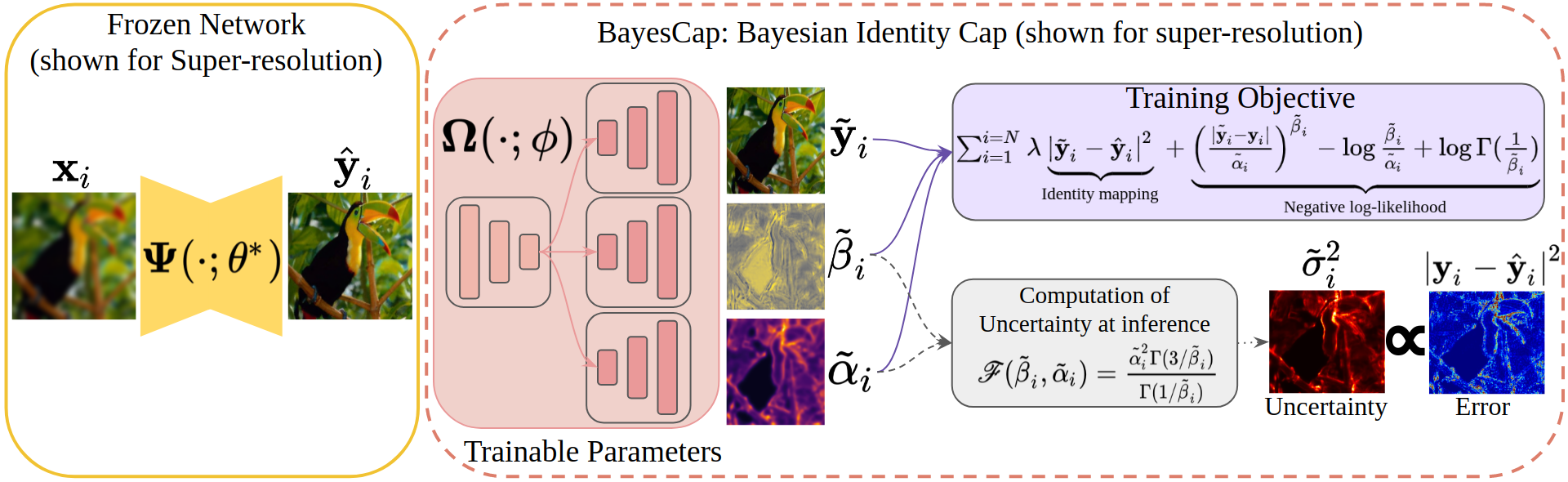}
    \vspace{-20pt}
    \caption{
        \texttt{BayesCap} ($\mathbf{\Omega}(\cdot; \phi)$) in tandem with the pretrained network with frozen parameters ($\mathbf{\Psi}(\cdot; \theta^*)$) (details in Section~\ref{sec:bayescap}). While the pretrained network cannot estimate the uncertainty, the proposed \texttt{BayesCap} feeds on the output of the pretrained network and maps it to the underlying probability distribution that allows computation of well calibrated uncertainty estimates.
    }
    \label{fig:arch}
    \vspace{-10pt}
\end{figure}
While the above formulation (Eq.~\eqref{eq:scratch_gauss}-\eqref{eq:sratch_ggd}) shows the dependence of various predicted distribution parameters on one another when maximizing the likelihood, it requires training the model from scratch, that we want to avoid. In the following, we describe how we address this problem through our \texttt{BayesCap}.

\subsection{Constructing \texttt{BayesCap}}
\label{sec:bayescap}
In the above, $\mathbf{\Psi}_s(\cdot; \zeta)$ was trained from scratch to predict all the parameters of distribution and does \textit{not} leverage the \textit{frozen} model $\mathbf{\Psi}(\cdot; \theta^*)$ estimating $\mathbf{y}_i$ using $\hat{\mathbf{y}}_i$ in a deterministic fashion.
To circumvent the training from scratch, we notice that one only needs to estimate the remaining parameters of the underlying distribution.
Therefore, to augment the frozen point estimation model 
, we learn a Bayesian identity mapping represented by $\mathbf{\Omega}(\cdot; \phi): \mathbb{R}^n \rightarrow \mathbb{R}^n$, that reconstructs the output of the frozen model $\mathbf{\Psi}(\cdot; \theta^*)$ and also produces the parameters of the distribution modeling the reconstructed output. We refer to this network as \texttt{BayesCap} (schematic in Figure~\ref{fig:arch}). As in Eq.~\eqref{eq:sratch_ggd}, we use heteroscedastic generalized Gaussian to model output distribution, i.e.,
\begin{gather}
\mathbf{\Omega}(\hat{\mathbf{y}}_i = \mathbf{\Psi}(\mathbf{x}_i; \theta^*); \phi) = \{\tilde{\mathbf{y}}_i, \tilde{\alpha}_i, \tilde{\beta}_i\} 
\text{, with } 
\mathbf{y}_i \sim \frac{\tilde{\beta}_i}{2 \tilde{\alpha}_i \Gamma(\frac{1}{\tilde{\beta}_i})} 
e^{-(|\tilde{\mathbf{y}}_i - \mathbf{y}_i|/\tilde{\alpha}_i)^{\tilde{\beta}_i}} 
\label{eq:omega}
\end{gather}
To enforce the identity mapping,
for every input $\mathbf{x}_i$, we regress the reconstructed output of the \texttt{BayesCap} ($\tilde{\mathbf{y}}_i$) with the output of the pretrained base network ($\hat{\mathbf{y}}_i$). 
This ensures that, the distribution predicted by \texttt{BayesCap} for an input $\mathbf{x}_i$, i.e., $\mathbf{\Omega}(\mathbf{\Psi}(\mathbf{x}_i; \theta^*); \phi)$, is such that the point estimates $\tilde{\mathbf{y}}_i$ match the point estimates of the pretrained network $\hat{\mathbf{y}}_i$.
Therefore, as the quality of the reconstructed output improves, the uncertainty estimated by $\mathbf{\Omega}(\cdot; \phi)$ also approximates the uncertainty for the prediction made by the pretrained $\mathbf{\Psi}(\cdot; \theta^*)$, i.e.,
\begin{gather}
    \tilde{\mathbf{y}}_i \rightarrow \hat{\mathbf{y}}_i \implies \tilde{\mathbf{\sigma}}_i^2 = \frac{\tilde{\alpha}_i^2 \Gamma(3/\tilde{\beta}_i)}{\Gamma(1/\tilde{\beta}_i)} \rightarrow \hat{\mathbf{\sigma}}_i^2 
\end{gather}

To train $\mathbf{\Omega}(\cdot; \phi)$ and obtain optimal parameters ($\phi^*$), we minimize the fidelity term between $\tilde{\mathbf{y}}_i$ and $\hat{\mathbf{y}}_i$, along with the negative log-likelihood for $\mathbf{\Omega}(\cdot; \phi)$, i.e.,
\begin{gather}
\phi^*
= \underset{\phi}{\text{argmin}} \sum_{i=1}^{N} \lambda \underbrace{|\tilde{\mathbf{y}}_i - \hat{\mathbf{y}}_i|^2}_{\text{Identity mapping}}
+
\underbrace{
\left(
\frac{|\tilde{\mathbf{y}}_{i}-\mathbf{y}_{i}|}{\tilde{\alpha}_{i}} \right)^{\tilde{\beta}_{i}} - 
    \log\frac{\tilde{\beta}_{i}}{\tilde{\alpha}_{i}} + \log\Gamma(\frac{1}{\tilde{\beta}_{i}})
}_{\text{Negative log-likelihood}}
\label{eq:bayescap}
\end{gather}
Here $\lambda$ represents the hyperparameter controlling the contribution of the fidelity term in the overall loss function. Extremely high $\lambda$ will lead to improper estimation of the ($\tilde{\alpha}$) and ($\tilde{\beta}$) parameters as other terms are ignored. Eq.~\eqref{eq:bayescap} allows \texttt{BayesCap} to estimate the underlying distribution and uncertainty.

Eq.~\eqref{eq:omega} and \eqref{eq:bayescap} show that the construction of $\mathbf{\Omega}$ is independent of $\mathbf{\Psi}$ and the $\mathbf{\Omega}$ always performs the Bayesian identity mapping regardless of the task performed by $\mathbf{\Psi}$. This suggests that task specific tuning will have minimal impact on the performance of $\mathbf{\Omega}$.
In our experiments, we employed the same network architecture for \texttt{BayesCap} with the same set of hyperparameters, across different tasks and it achieves state-of-the-art uncertainty estimation results without task specific tuning (as shown in Section~\ref{sec:exp}), highlighting
that \texttt{BayesCap} is not sensitive towards various design choices including architecture, learning-rate, etc. %

\section{Experiments}
\label{sec:exp}
\vspace{-5pt}
We first describe our experimental setup (i.e., datasets, tasks, and evaluation metrics) in Section \ref{sec:setup}. We compare our model to a wide variety of state-of-the-art methods quantitatively and qualitatively in Section \ref{sec:sota}.
Finally in Section \ref{sec:ablation}, we provide an ablation analysis along with a real world application of \texttt{BayesCap} for detecting out-of-distribution samples. %

\subsection{Tasks and Datasets}
\label{sec:setup}
We show the efficacy of our \texttt{BayesCap} on various image enhancement and translation tasks including super-resolution, deblurring, inpainting, and MRI translation, as detailed below. In general, image enhancement and translation tasks are highly ill-posed problems as an \textit{injective function} between input and output may not exist~\cite{liu2017unsupervised,upadhyay2021robustness}, thereby necessitating the need to learn a probabilistic mapping to quantify the uncertainty and indicate poor reconstruction of the output images.
For each task we choose a well established deterministic pretrained network, for which we estimate uncertainties.

\textbf{Super-resolution.}
The goal is to map low-resolution images to their %
high-resolution counterpart.
We choose pretrained SRGAN~\cite{Ledig_2017_CVPR} as our base model $\mathbf{\Psi}(\cdot; \theta^*)$. The \texttt{BayesCap} model $\mathbf{\Omega}(\cdot; \phi)$ is trained on ImageNet patches sized $84\times84$ to perform $4\times$ super-resolution. The resulting combination of SRGAN and \texttt{BayesCap} is evaluated on the Set5~\cite{set5}, Set14~\cite{set14}, and BSD100~\cite{bsd100} datasets. 

\textbf{Deblurring.} 
The goal is to remove noise from images corrupted with blind motion. %
We use the pretrained DeblurGANv2~\cite{kupyn2019deblurgan} which shows improvements over the original DeblurGAN~\cite{kupyn2018deblurgan}. The \texttt{BayesCap} model is evaluated on the %
GoPro dataset~\cite{nah2017deep}, using standard train/test splits. %

\textbf{Inpainting.} 
The goal is to fill masked regions of an input image. %
We use pretrained DeepFillv2~\cite{yu2019free}, that improves over DeepFill~\cite{yu2018generative}, as the base model for inpainting. Both the original base model and the \texttt{BayesCap} are trained and tested on the standard train/test split of Places365 dataset~\cite{zhou2017places}.

\textbf{MRI Translation.}
We predict the T2 MRI from T1 MRI, an important problem in medical imaging as discussed in~\cite{chartsias2017multimodal,bowles2016pseudo,iglesias2013synthesizing,ye2013modality,yu2019ea}. 
We use the pretrained deterministic UNet as base model~\cite{upadhyay2021uncertaintyICCV,unet}. Both the base model and \texttt{BayesCap} are trained and tested on IXI datatset~\cite{robinson2010identifying} following~\cite{upadhyay2021uncertaintyICCV}.

\textbf{Baselines.} For all tasks, we compare \texttt{BayesCap} against 7 methods in total, out of which 6 baselines can estimate uncertainty of a pretrained model without re-training and one baseline modifies the base network and train it from \textit{scratch} to estimate the uncertainty.  %
The first set of baselines belong to \textit{test-time data augmentation} (\texttt{TTDA}) technique~\cite{ayhan2018test,wang2018test,wang2019aleatoric}, where we generate multiple perturbed copies of the input and use the set of corresponding outputs to compute the uncertainty. 
We consider three different ways of perturbing the input, (i)~per-pixel noise perturbations \texttt{TTDAp}~\cite{ayhan2018test,wang2018test}, (ii)~affine transformations \texttt{TTDAa}~\cite{ayhan2018test,wang2018test} and (iii)~random corruptions from Gaussian blurring, contrast enhancement, and color jittering (\texttt{TTDAc})~\cite{ayhan2018test,wang2018test}. %
As additional baseline, we also consider \texttt{TTDApac} that generates the multiple copies by combining pixel-level perturbations, affine transformations, and corruptions as described above. 

Another set of baselines uses dropout~\cite{srivastava2014dropout,kendall2015bayesian,gal2016dropout,laves2019well,maddox2019simple} before the final predictions. This is possible even for the models \textit{that are not originally trained with dropout}. We refer to this model as \texttt{DO}. In addition, we consider a baseline that combines dropout with test-time data augmentation (\texttt{DOpac}). 
Finally, we also compare against a model trained from scratch to produce the uncertainty as described in~\cite{kendall2017uncertainties}. We refer to this as \texttt{Scratch}. %

\textbf{Metrics.} 
We evaluate the performance of various models on two kinds of metrics (i)~image reconstruction quality and (ii)~predicted uncertainty calibration quality.
To measure reconstruction quality, we use SSIM~\cite{wang2004image} and PSNR. For inpainting we also show mean $\ell_1$ and $\ell_2$ error, following the convention in original works~\cite{yu2018generative,yu2019free}. 
We emphasize that all the methods, \textit{except} \texttt{Scratch}, can use the output of the pretrained base model and only derive the uncertainty maps using different estimation techniques described above. Therefore image reconstruction quality metrics like SSIM, PSNR, mean $\ell_1$ and $\ell_2$ error remain the same as that of base network. However, \texttt{Scratch} method \textit{does not} have access to the pretrained model, therefore it has to use its own predicted output and uncertainty estimates. 

To quantify the quality of the uncertainty, we use the \textit{uncertainty calibration error} (UCE) as described in~\cite{laves2020calibration,guo2017calibration} for regression tasks. It measures the discrepancy between the predictive error and predictive uncertainty, given by, 
$\text{\texttt{UCE}} := \sum_{m=1}^M \nicefrac{|B_m|}{N} |\text{\texttt{err}}(B_m) - \text{\texttt{uncer}}(B_m)|$, where
$B_m$ is one of the uniformly separated bins,
$\text{\texttt{err}}(B_m) := \nicefrac{1}{|B_m|} \sum_{i \in B_m} ||\hat{\mathbf{y}}_i - \mathbf{y}_i||^2$, and $\text{\texttt{uncer}}(B_m) := \nicefrac{1}{|B_m|} \sum_{i \in B_m} \hat{\mathbf{\sigma}}_i^2$.
We also use \textit{correlation coefficient} (C.Coeff.) between the error and the uncertainty, as high correlation is desirable.

{\bf Implementation Details.} We optimize Eq.~\eqref{eq:bayescap} using the Adam optimizer~\cite{kingma2014adam} and a batch size of 2 with images that are resized to $256\times256$. During training we exponentially anneal the hyperparameter $\lambda$ that is initially set to 10. This guides the \texttt{BayesCap} to learn the identity mapping in the beginning, and gradually learn the optimal parameters of the underlying distribution via maximum likelihood estimation. %

\begin{figure}[!t]
    \centering
    \includegraphics[width=\textwidth]{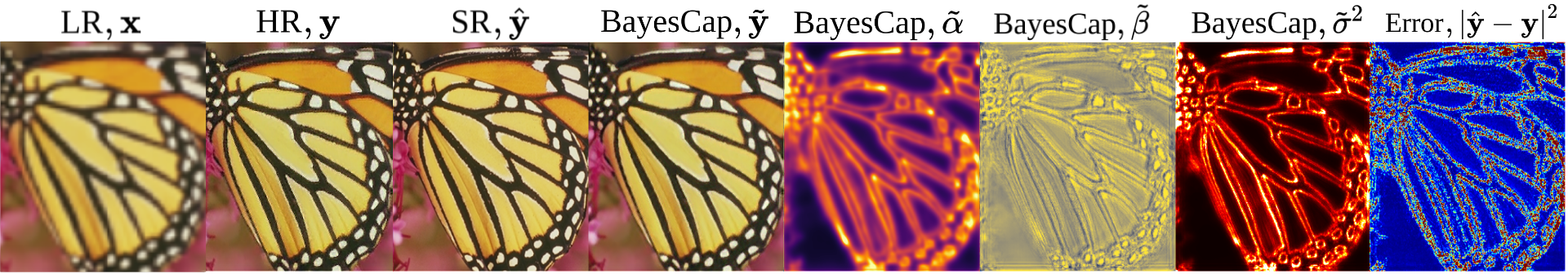}
    \vspace{-20pt}
    \caption{
        Input (LR,$\mathbf{x}$) and output of pretrained SRGAN (SR,$\hat{\mathbf{y}}$) along with output of  \texttt{BayesCap} ($\{\tilde{\mathbf{y}}, \tilde{\mathbf{\alpha}}, \tilde{\mathbf{\beta}}\}$).
        Spatially varying parameters ($\tilde{\alpha}, \tilde{\beta}$) lead to well-calibrated uncertainty $\tilde{\sigma}^2$, highly correlated with the SRGAN error, $|\mathbf{y}-\hat{\mathbf{y}}|^2$.
    }
    \label{fig:ta}
    \vspace{3pt}
    \centering
    \includegraphics[width=\textwidth]{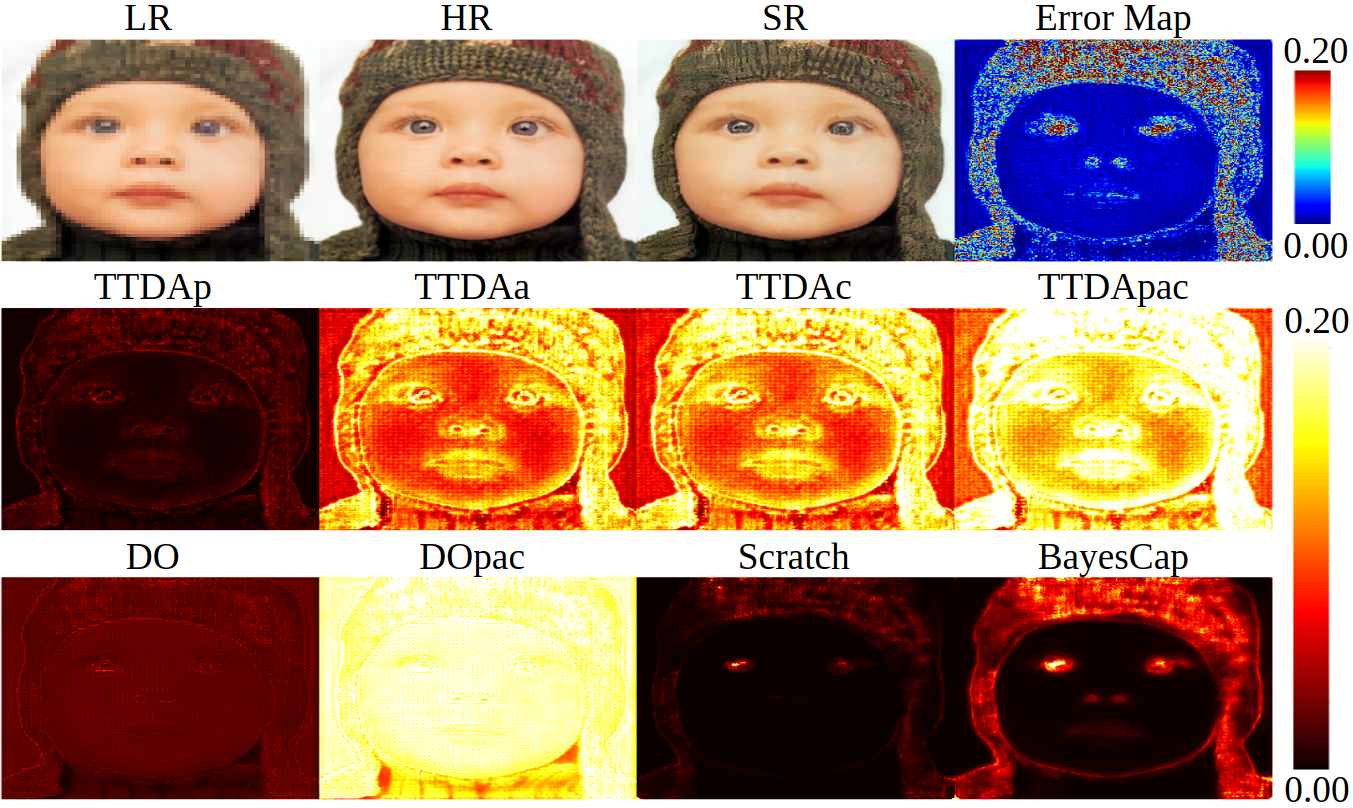}
    \vspace{-20pt}
    \caption{
        Qualitative example showing the results of the pre-trained SRGAN model along with the uncertainty maps produced by \texttt{BayesCap} and the other methods.
        Uncertainty derived from \texttt{BayesCap} has better correlation with the error.
    }
    \label{fig:srgan}
    \vspace{-10pt}
\end{figure}

\begin{figure}[t]
{
\setlength{\tabcolsep}{2pt}
\renewcommand{\arraystretch}{1.2}
\resizebox{\linewidth}{!}{
\begin{tabular}{l|c|a|bbbb|bb|e|d}
  \multirow{1}{*}{\textbf{D}} & \multirow{1}{*}{\textbf{Metrics}} & \textbf{SRGAN} & \textbf{TTDAp} & \textbf{TTDAa} & \textbf{TTDAc} & \textbf{TTDApac} & \textbf{DO} & \textbf{DOpac} & \textbf{Scratch} & \textbf{BayesCap} \\
  \hline
  \hline
  \multirow{4}{*}{\rotatebox[origin=c]{90}{\textbf{Set5}}}
  & PSNR$\uparrow$ & 29.40 & 29.40 & 29.40 & 29.40 & 29.40 & 29.40 & 29.40 & 27.83 & 29.40 \\
  & SSIM$\uparrow$ & 0.8472 & 0.8472 & 0.8472 & 0.8472 & 0.8472 & 0.8472 & 0.8472 & 0.8166 & 0.8472 \\
  & UCE$\downarrow$ & \textbf{NA} & 0.39 & 0.40 & 0.42 & 0.47 & 0.33 & 0.36 & 0.035 & \textbf{0.014} \\
  & C.Coeff$\uparrow$ & \textbf{NA} & 0.17 & 0.13 & 0.08 & 0.03 & 0.05 & 0.07 & 0.28 & \textbf{0.47} \\
  \hline
  \multirow{4}{*}{\rotatebox[origin=c]{90}{\textbf{Set14}}}
  & PSNR$\uparrow$ & 26.02 & 26.02 & 26.02 & 26.02 & 26.02 & 26.02 & 26.02 & 25.31 & 26.02 \\
  & SSIM$\uparrow$ & 0.7397 & 0.7397 & 0.7397 & 0.7397 & 0.7397 & 0.7397 & 0.7397 & 0.7162 & 0.7397 \\
  & UCE$\downarrow$ & \textbf{NA} & 0.57 & 0.63 & 0.61 & 0.69 & 0.48 & 0.52 & 0.048 & \textbf{0.017} \\
  & C.Coeff$\uparrow$ & \textbf{NA} & 0.07 & 0.04 & 0.04 & 0.06 & 0.08 & 0.04 & 0.22 & \textbf{0.42} \\
  \hline
  \multirow{4}{*}{\rotatebox[origin=c]{90}{\textbf{BSD100}}}
  & PSNR$\uparrow$ & 25.16 & 25.16 & 25.16 & 25.16 & 25.16 & 25.16 & 25.16 & 24.39 & 25.16 \\
  & SSIM$\uparrow$ & 0.6688 & 0.6688 & 0.6688 & 0.6688 & 0.6688 & 0.6688 & 0.6688 & 0.6297 & 0.6688 \\
  & UCE$\downarrow$ & \textbf{NA} & 0.72 & 0.77 & 0.81 & 0.83 & 0.61 & 0.64 & 0.057 & \textbf{0.028} \\
  & C.Coeff$\uparrow$ & \textbf{NA} & 0.13 & 0.09 & 0.11 & 0.09 & 0.10 & 0.08 & 0.31 & \textbf{0.45} \\
  \hline
\end{tabular}%
}
\vspace{-10pt}
\captionof{table}{
Quantitative results showing the performance of pretrained SRGAN in terms of PSNR and SSIM, along with the quality of of uncertaintiy maps obtained by \texttt{BayesCap} and other baselines, in terms of UCE and Correlation Coefficient (C.Coeff). All results on 3 datasets including Set5, Set14, and BSD100.
}
\label{tab:quantsrgan}
}
\vspace{-15pt}
\end{figure}

\subsection{Results}
\label{sec:sota}
\vspace{-4pt}
\textbf{Super-resolution.}
Table~\ref{tab:quantsrgan} shows the image reconstruction performance along with the uncertainty calibration performance on the Set5, Set14, and the BSD100 datasets for all the methods.
We see that \texttt{BayesCap} significantly outperforms all other methods in terms of UCE and C.Coeff while retaining the image reconstruction performance of the base model. For instance, across all the 3 sets, the correlation coefficient is always greater than 0.4 showing that the error and the uncertainty are correlated to a high degree. The model trained from scratch has a correlation coefficient between 0.22-0.31 which is lower than \texttt{BayesCap}. The baselines based on dropout and test-time data augmentation show nearly no correlation between the uncertainty estimates and the error (C.Coeff. of 0.03-0.17). A similar trend can be seen with the UCE where \texttt{BayesCap} has the best UCE scores followed by the model trained from scratch, while the test-time data augmentation and the dropout baselines have a very high UCE (between 0.33 and 0.83) suggesting poorly calibrated uncertainty estimates. 

Qualitatively, Figure~\ref{fig:ta} shows the prediction of the pretrained SRGAN along with the predictions of the \texttt{BayesCap} showing per-pixel estimated distribution parameters along with the uncertainty map on a sample from Set5 dataset. High correlation between the per-pixel predictive error of SRGAN and uncertainties from \texttt{BayesCap} suggests that \texttt{BayesCap} produces well-calibrated uncertainty estimates. Moreover, Figure~\ref{fig:srgan} shows that uncertainty produced by other baselines are not in agreement with the error (e.g., \texttt{TTDAp} does not show high uncertainty within the eye, where error is high) indicating that they are poorly calibrated.

\begin{figure}[!t]
    \centering
    \includegraphics[width=\textwidth,trim=0 38 0 0, clip]{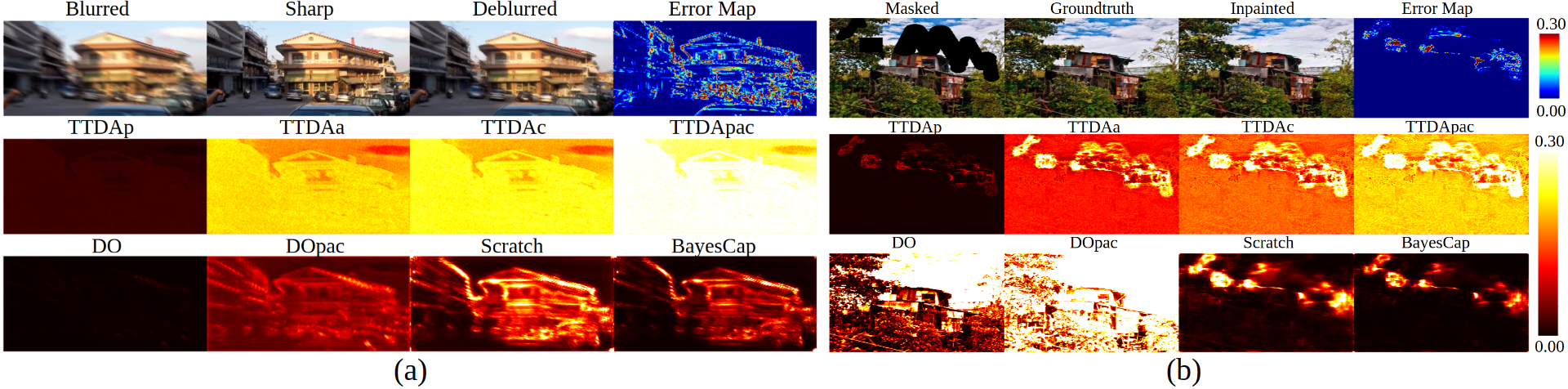}
    \vspace{-20pt}
    \caption{
       Qualitative example showing the results of the pretrained DeblurGANv2 and DeepFillv2 on image deblurring (left) and inpainting (right) tasks along with the uncertainty maps produced by different methods. %
    }
    \label{fig:deblur}
    
    \vspace{5pt}
    {
    \setlength{\tabcolsep}{2pt}
    \renewcommand{\arraystretch}{1.2}
    \resizebox{\linewidth}{!}{
    \begin{tabular}{l|c|a|bbbb|bb|e|d}
      \multirow{1}{*}{\textbf{D}} & \multirow{1}{*}{\textbf{Metrics}} & \textbf{DeblurGANv2} & \textbf{TTDAp} & \textbf{TTDAa} & \textbf{TTDAc} & \textbf{TTDApac} & \textbf{DO} & \textbf{DOpac} & \textbf{Scratch} & \textbf{BayesCap} \\
      \hline
      \hline
      \multirow{4}{*}{\rotatebox[origin=c]{90}{\textbf{GoPro}}}
      & PSNR$\uparrow$ & 29.55 & 29.55 & 29.55 & 29.55 & 29.55 & 29.55 & 29.55 & 26.16 & 29.55 \\
      & SSIM$\uparrow$ & 0.9340 & 0.9340 & 0.9340 & 0.9340 & 0.9340 & 0.9340 & 0.9340 & 0.8136 & 0.9340 \\
      & UCE$\downarrow$ & \textbf{NA} & 0.44 & 0.45 & 0.49 & 0.53 & 0.52 & 0.59 & 0.076 & \textbf{0.038} \\
      & C.Coeff$\uparrow$ & \textbf{NA} & 0.17 & 0.13 & 0.08 & 0.03 & 0.05 & 0.07 & 0.21 & \textbf{0.32} \\
      \hline
      \end{tabular}%
    }
    \vspace{-10pt}
    \captionof{table}{
    Results showing the performance of pretrained DeblurGANv2 in terms of PSNR and SSIM, along with the quality of of uncertaintiy obtained by \texttt{BayesCap} and other methods, in terms of UCE and C.Coeff on GoPro dataset. 
    }
    \label{tab:quantdeblur}
    }
    
    \vspace{5pt}
    {
    \setlength{\tabcolsep}{2pt}
    \renewcommand{\arraystretch}{1.2}
    \resizebox{\linewidth}{!}{
    \begin{tabular}{l|c|a|bbbb|bb|e|d}
      \multirow{1}{*}{\textbf{D}} & \multirow{1}{*}{\textbf{Metrics}} & \textbf{DeepFillv2} & \textbf{TTDAp} & \textbf{TTDAa} & \textbf{TTDAc} & \textbf{TTDApac} & \textbf{DO} & \textbf{DOpac} & \textbf{Scratch} & \textbf{BayesCap} \\
      \hline
      \hline
      \multirow{6}{*}{\rotatebox[origin=c]{90}{\textbf{Places365}}}
      & m. $\ell_1$ err.$\downarrow$ & 9.1\% & 9.1\% & 9.1\% & 9.1\% & 9.1\% & 9.1\% & 9.1\% & 15.7\% & 9.1\% \\
      & m. $\ell_2$ err.$\downarrow$ & 1.6\% & 1.6\% & 1.6\% & 1.6\% & 1.6\% & 1.6\% & 1.6\% & 5.8\% & 1.6\% \\
      & PSNR$\uparrow$ & 18.34 & 18.34 & 18.34 & 18.34 & 18.34 & 18.34 & 18.34 & 17.24 & 18.34 \\
      & SSIM$\uparrow$ & 0.6285 & 0.6285 & 0.6285 & 0.6285 & 0.6285 & 0.6285 & 0.6285 & 0.6032 & 0.6285 \\
      & UCE$\downarrow$ & \textbf{NA} & 0.63 & 0.88 & 0.87 & 0.93 & 1.62 & 1.49 & 0.059 & \textbf{0.011} \\
      & C.Coeff$\uparrow$ & \textbf{NA} & 0.26 & 0.11 & 0.12 & 0.08 & 0.09 & 0.12 & 0.44 & \textbf{0.68} \\
      \hline
      \end{tabular}%
    }
    \vspace{-10pt}
    \captionof{table}{
    Performance of pretrained DeepFillv2 in terms of mean $\ell_1$ error, mean $\ell_2$ error, PSNR and SSIM, along with the quality of uncertainty obtained by \texttt{BayesCap} and other methods, in terms of UCE and C.Coeff on Places365 dataset. 
    }
    \label{tab:quantpaint}
    }
     \vspace{-15pt}
\end{figure}

\textbf{Deblurring.} 
We report the results on the GoPro dataset in Table~\ref{tab:quantdeblur}. DeblurGANv2 achieves significantly better results than \texttt{Scratch} (29.55 vs 26.16 PSNR). In terms of UCE, \texttt{BayesCap} outperforms all the methods by achieving a low score of 0.038. While the \texttt{Scratch} is close and achieves a UCE of 0.076, all the other methods have a UCE that is nearly 10 times higher suggesting that \texttt{BayesCap} estimates the most calibrated uncertainty. This is also visible in Figure~\ref{fig:deblur}-(left) where the uncertainties provided by \texttt{BayesCap} is correlated with the error (C.Coeff. of 0.32) unlike methods from \texttt{TTDA} and \texttt{DO} class that have very low correlation between the uncertainty and the error (C.Coeff of 0.03 - 0.17). While \texttt{Scratch} achieves a reasonable score, second only to \texttt{BayesCap}, in terms of UCE (0.076 vs. 0,038) and C.Coeff (0.21 vs. 0.32), it has much poorer image reconstruction output with a PSNR of 26.16 and SSIM of 0.8136. The poor reconstruction also justifies the relatively higher uncertainty values for \texttt{Scratch} (as seen in Figure~\ref{fig:deblur}-(left)) when compared to \texttt{BayesCap}.

\textbf{Inpainting.} Table~\ref{tab:quantpaint} shows the results on Places365 dataset. The pretrained base model DeepFillv2~\cite{yu2019free} achieves a mean L1 error of 9.1\%, however \texttt{Scratch} is much worse, achieving a mean L1 error of 15.7\%. This again demonstrates that training a Bayesian model from scratch often does not replicate the performance of deterministic counterparts. 
Also, \texttt{BayesCap} retains the reconstruction performance of DeepFillv2~\cite{yu2019free} and provides well-calibrated uncertainties, as demonstrated by highest C.Coeff. (0.68), and the lowest UCE (0.011). Methods belonging to \texttt{TTDA} and \texttt{DO} classes are unable to provide good uncertainties (C.Coeff. of 0.08-0.26). %
The example in Figure~\ref{fig:deblur}-(right) also illustrates an interesting phenomenon. Although, the uncertainties are predicted for the entire image, we see that \texttt{BayesCap} automatically learns to have extremely low uncertainty values outside the masked region which is perfectly reconstructed. Within the masked region, uncertainty estimates are highly correlated with the error. 

\begin{figure}[!t]
    \centering
    \includegraphics[width=\textwidth, height=0.5\textwidth]{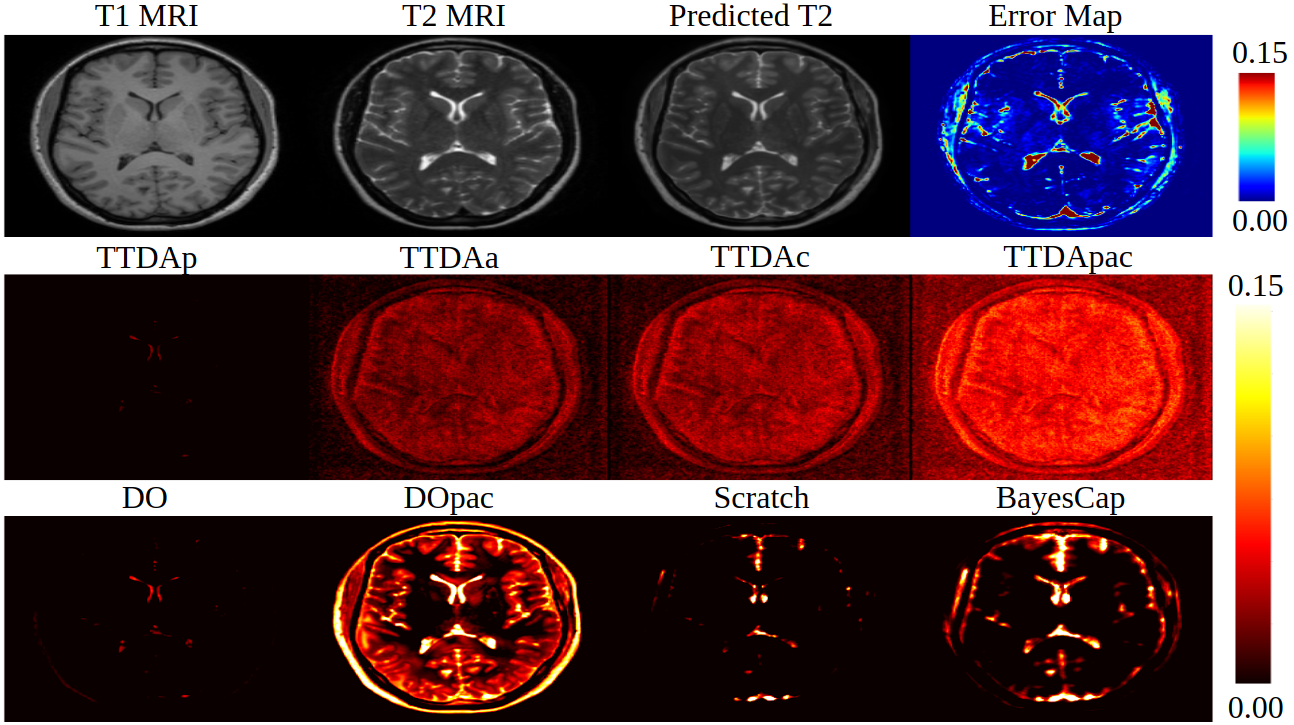}
    \vspace{-20pt}
    \caption{
        Qualitative example showing the results of the pretrained UNet for T1 to T2 MRI translation along with the uncertainty produced by different methods.
    }
    \label{fig:mri}
\vspace{5pt}
{
\setlength{\tabcolsep}{2pt}
\renewcommand{\arraystretch}{1.2}
\resizebox{\linewidth}{!}{
\begin{tabular}{l|c|a|bbbb|bb|e|d}
  \multirow{1}{*}{\textbf{D}} & \multirow{1}{*}{\textbf{Metrics}} & \textbf{UNet} & \textbf{TTDAp} & \textbf{TTDAa} & \textbf{TTDAc} & \textbf{TTDApac} & \textbf{DO} & \textbf{DOpac} & \textbf{Scratch} & \textbf{BayesCap} \\
  \hline
  \hline
  \multirow{4}{*}{\rotatebox[origin=c]{90}{\textbf{IXI}}}
  & PSNR$\uparrow$ & 25.70 & 25.70 & 25.70 & 25.70 & 25.70 & 25.70 & 25.70 & 25.50 & 25.70 \\
  & SSIM$\uparrow$ & 0.9272 & 0.9272 & 0.9272 & 0.9272 & 0.9272 & 0.9272 & 0.9272 & 0.9169 & 0.9272 \\
  & UCE$\downarrow$ & \textbf{NA} & 0.53 & 0.46 & 0.41 & 0.44 & 0.38 & 0.40 & 0.036 & \textbf{0.029} \\
  & C.Coeff$\uparrow$ & \textbf{NA} & 0.05 & 0.14 & 0.16 & 0.08 & 0.13 & 0.47 & 0.52 & \textbf{0.58} \\
  \hline
  \end{tabular}%
}
\vspace{-10pt}
\captionof{table}{
Performance of pretrained UNet for MRI translation in terms of PSNR and SSIM, along with the quality of of uncertainty obtained by \texttt{BayesCap} and other methods, in terms of UCE and C.Coeff on IXI Dataset.
}
\label{tab:quantmri}
}
\vspace{-15pt}
\end{figure}

\textbf{MRI Translation.}
We perform T1 to T2 MRI translation as described in~\cite{upadhyay2021uncertaintyICCV,upadhyay2021robustness} which has an impact in clinical settings by reducing MRI acquisition times~\cite{chartsias2017multimodal,bowles2016pseudo,iglesias2013synthesizing,ye2013modality,yu2019ea}.
The quantitative results on the IXI dataset are show in Table~\ref{tab:quantmri}. The pretrained base model employing U-Net architecture, achieves a SSIM score of 0.9272 which is nearly matched by \texttt{Scratch} (0.9169). However, we see that \texttt{BayesCap} performs better than \texttt{Scratch} in terms of UCE (0.036 vs. 0.029) and C.Coeff (0.52 vs. 0.58). Other methods are poorly calibrated, as indicated by high UCE and low C.Coeff. This is also evident from Figure~\ref{fig:mri}, indicating high correlation between the \texttt{BayesCap} uncertainty and error, while low correlation of the same for other methods.

\begin{figure}[t]
    \centering
    \includegraphics[width=\textwidth]{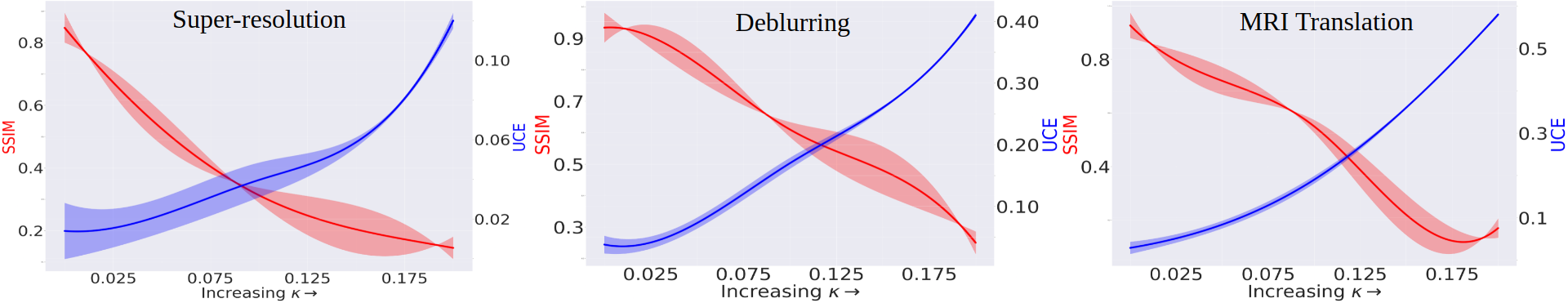}
    \vspace{-17pt}
    \caption{
        \textit{Impact of the identity mapping.}  %
        Degrading the quality of the identity mapping (\textcolor{red}{SSIM}) at inference, leads to poorly calibrated uncertainty (\textcolor{blue}{UCE}). $\kappa$ represents the magnitude of noise used for degrading the identity mapping.
    }
    \label{fig:idg}

    \includegraphics[width=\textwidth]{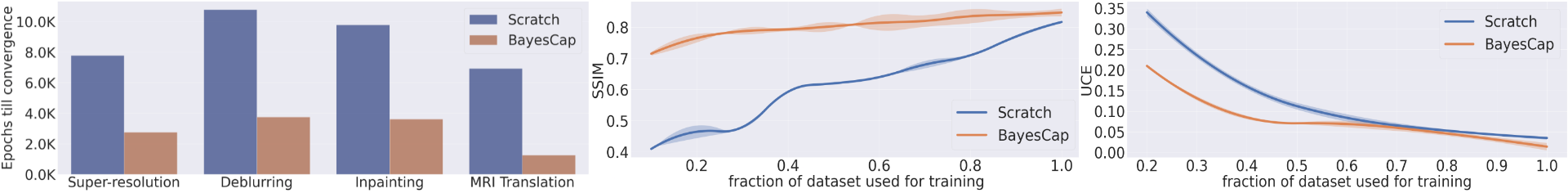}
    \vspace{-17pt}
    \caption{ 
        \texttt{BayesCap} can be trained to achieve optimal performance in fewer epochs (left), while being more data-efficient (achieves better results with fewer samples) as compared to \texttt{Scratch} (middle and right), shown for super-resolution. 
    }
    \label{fig:footprint}
    \vspace{-10pt}
\end{figure}

\vspace{-10pt}
\subsection{Ablation Studies}
\label{sec:ablation}
\vspace{-4pt}

As discussed in Section~\ref{sec:bayescap}, \texttt{BayesCap} can help in estimating uncertainty for the frozen model only if \texttt{BayesCap} provides perfect reconstruction.
To study this, we design an experiment that deteriorates the identity mapping learned by \texttt{BayesCap}, leading to poor uncertainty estimates. Moreover, we also demonstrate that \texttt{BayesCap} is much more data and compute efficient when compared to \texttt{Scratch} (i.e., model capable of estimating the uncertainty, trained from scratch).

Figure~\ref{fig:idg} shows that preserving the identity mapping is essential to providing well-calibrated post-hoc uncertainty estimates using \texttt{BayesCap}. Here, we gradually 
feed increasingly noisy samples (corrupted by zero-mean Gaussian with variance given by $\kappa^2$) to the model that leads to poor reconstruction by \texttt{BayesCap} and as a result degrades the identity mapping.
As the quality of the identity mapping degrades (decreasing SSIM), we see that quality of uncertainty also degrades (increasing UCE). For instance, for super-resolution, with zero noise in the input the reconstruction quality of the \texttt{BayesCap} is at its peak (SSIM, 0.8472) leading to almost identical mapping. This results in well-calibrated %
uncertainty maps derived from \texttt{BayesCap} %
(UCE, 0.014). However, with $\kappa=0.15$, the reconstruction quality decreases sharply (SSIM of 0.204) leading to poorly calibrated uncertainty (UCE of 0.0587), justifying the need for the identity mapping.

We also show that
\texttt{BayesCap} is more efficient than training a Bayesian model from scratch both in terms of time and data required to train the model in Figure~\ref{fig:footprint}-(Left). On all the 4 tasks, \texttt{BayesCap} can be trained 3-5$\times$ faster than \texttt{Scratch}. For super-resolution, we show that \texttt{BayesCap} can achieve competitive results even when trained on a fraction of the dataset, whereas the performance of \texttt{Scratch} reduces sharply in low data regime, as shown in Figure~\ref{fig:footprint}-(Middle and Right). For instance, with just 50\% of training data, \texttt{BayesCap} performs 33\% better than \texttt{Scratch} in terms of SSIM. This is because \texttt{BayesCap} learns the autoencoding task, whereas \texttt{Scratch} learns the original translation task.

\subsection{Application: Out-of-Distribution Analysis}
\label{sec:ood}
\begin{figure}[t]
    \centering
    \includegraphics[width=\textwidth]{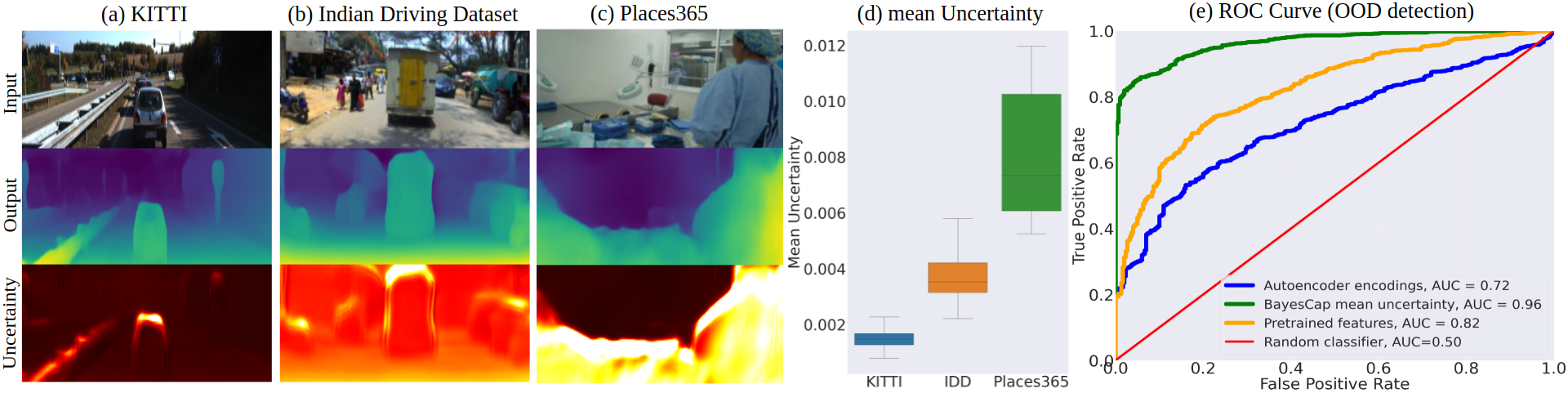}
    \vspace{-18pt}
    \caption{
        \texttt{BayesCap} with MonoDepth2~\cite{godard2019digging} for depth estimation in autonomous driving. Trained on KITTI and evaluated on (a) KITTI, (b) Indian Driving Dataset, and (c) Places365. (d) and (e) Plots show mean uncertainty values and ROC curve for OOD detection respectively, as described in Section~\ref{sec:ood}.
    }
    \label{fig:depth}
    \vspace{-10pt}
\end{figure}

The proposed \texttt{BayesCap} focuses on estimating well-calibrated uncertainties of pretrained image regression models in a post-hoc fashion. This can be crucial in many critical real-world scenarios such as autonomous driving and navigation~\cite{xu2014motion,kendall2017uncertainties,wang2019pseudo,michelmore2018evaluating}. We consider monocular depth estimation (essential for autonomous driving) and show that the estimated uncertainty can help in detecting \textit{out-of-distribution} (OOD) samples.
We take MonoDepth2~\cite{godard2019digging} that is trained on the KITTI dataset~\cite{geiger2013vision}. The resulting model  with \texttt{BayesCap} is evaluated on the validation sets of the KITTI dataset, the India Driving Dataset(IDD)~\cite{varma2019idd} as well as the Places365 dataset~\cite{zhou2017places}. The KITTI dataset captures images from German cities, while the India Driving Dataset has images from Indian roads. Places365 consists of images for scene recognition and is vastly different from driving datasets. 
Intuitively, both IDD and Places365 represent OOD as there is a shift in distribution when compared to training data (KITTI), this is captured in Figure~\ref{fig:depth}-(a,b,c,d), representing degraded depth and uncertainty on OOD images. The uncertainties also reflect this %
with increasingly higher values for OOD samples as also shown in the bar plot (Figure~\ref{fig:depth}-(d)). This suggests that mean uncertainty value for an image can be used to detect if it belongs to OOD.

To quantify this, we plot the ROC curve for OOD detection on a combination of the KITTI (in-distribution) samples and IDD and Places365 (out-of-distribution) samples in Figure~\ref{fig:depth}-(e). Additionally, we compare using uncertainty for OOD detection against (i)~using intermediate features from the pretrained model~\cite{reiss2021panda} and (ii)~using features from the bottleneck of a trained autoencoder~\cite{zhou2017anomaly}. Samples are marked OOD if the distance between the features of the samples and mean feature of the KITTI validation set is greater than a threshold. We clearly see that using the mean uncertainty achieves far superior results (AUROC of 0.96), against the pretrained features (AUROC of 0.82) and autoencoder based approach (AUROC of 0.72). 
Despite not being specifically tailored for OOD detection, \texttt{BayesCap} achieves strong results. This indicates the benefits of its well-calibrated uncertainty estimates on downstream tasks.

\vspace{-8pt}
\section{Conclusions}
\vspace{-8pt}
We proposed \texttt{BayesCap}, a fast and cheap post-hoc uncertainty estimation method for pretrained deterministic models. %
We show that our method consistently produces well-calibrated uncertainty estimates across a wide variety of image enhancement and translation tasks without hampering the performance of the pretrained model. This is in sharp contrast to training a Bayesian model from scratch that is more expensive and often not competitive with deterministic counterparts. We demonstrate that derived calibrated uncertainties can be used in critical scenarios for detecting OOD and helping decision-making. One limitation of our model is the assumption that %
the input is sufficiently contained in the output of the target task. Future works may address this limitation, as well as extending \texttt{BayesCap} to discrete predictions. %

\myparagraph{Acknowledgments.}
This work has been partially funded by the ERC (853489 - DEXIM), by the DFG (2064/1 – Project number 390727645). %
The authors thank the International Max Planck Research School for Intelligent Systems (IMPRS-IS) for supporting Uddeshya Upadhyay and Shyamgopal Karthik. 

\bibliographystyle{splncs04}
\bibliography{egbib}
\end{document}


\pagestyle{headings}
\mainmatter
\def\ECCVSubNumber{1824}  %

\title{BayesCap: Bayesian Identity Cap for Calibrated Uncertainty in Frozen Neural Networks \\ (Supplementary Material)} %

\titlerunning{\texttt{BayesCap}: Bayesian Identity Cap for Fast Calibrated Uncertainty}
\author{Uddeshya Upadhyay$^{*}$\inst{1} \and
Shyamgopal Karthik$^{*}$\inst{1} \and
Yanbei Chen,\inst{1} \\
Massimiliano Mancini\inst{1} \and
Zeynep Akata\inst{1,2}}
\authorrunning{U. Upadhyay et al.}
\institute{University of T\"{u}bingen \and
Max Planck Institute for Intelligent Systems}
\maketitle

We first provide additional baselines for the ablation study showing the importance of the identity mapping. Then, we discuss 
the experimental details for the out-of-distribution analysis presented in the main paper,
The code for the \texttt{BayesCap} is available at \url{https://github.com/ExplainableML/BayesCap}.

\sloppy

\section{More Ablation Studies}
\subsection{Identity Degradation.}
\begin{figure}[!h]
    \centering
    \includegraphics[width=0.99\textwidth]{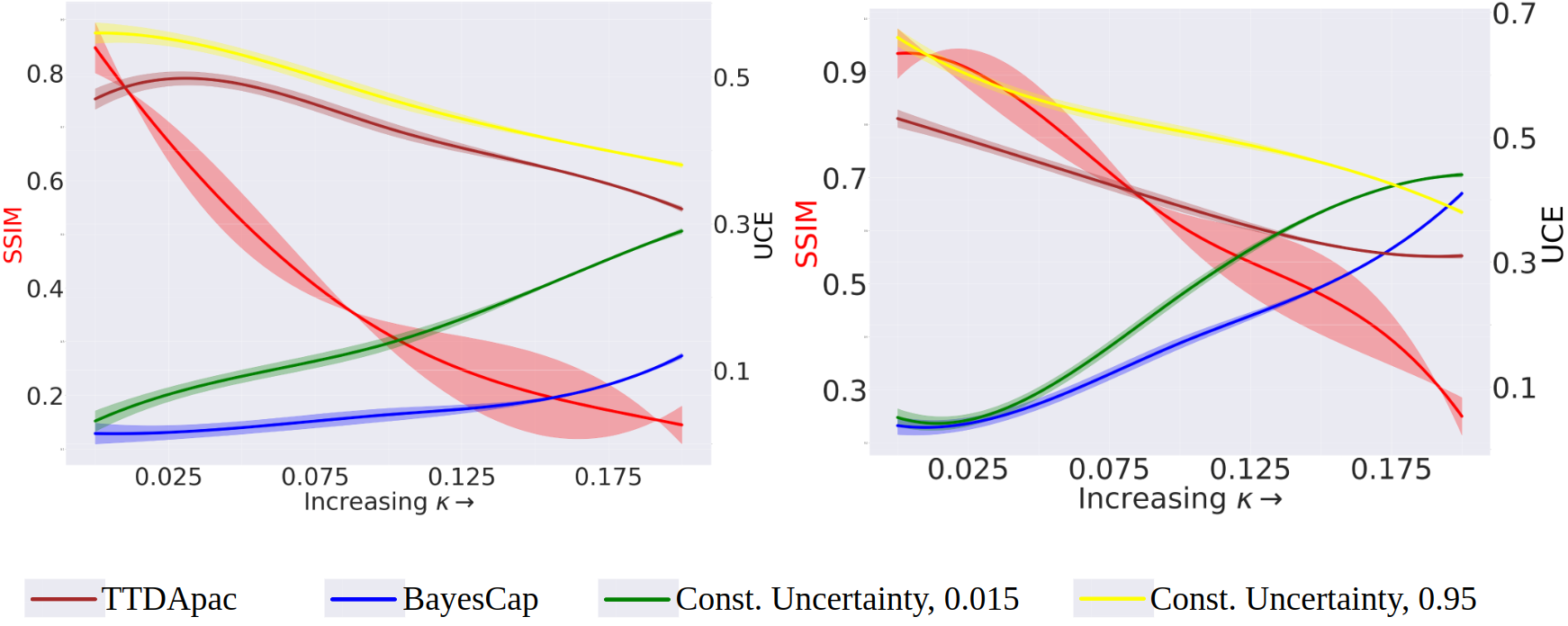}
    \caption{\textit{Impact of the identity mapping.} Degrading the quality of the identity mapping (SSIM) at inference, leads to poorly calibrated uncertainty (UCE). $\kappa$ represents the magnitude of noise used for degrading the identity mapping. (Left) super-resolution on set5 dataset and (Right) Deblurring on GoPro dataset.}
    \label{fig:id}
\end{figure}

We study the identity degradation performance for \texttt{TTDApac} just like we do for \texttt{BayesCap} in the main manuscript. The results of the experiments on super-resolution and deblurring task are shown in Figure~\ref{fig:id}.

We notice that, at the beginning when the input samples are not degraded the UCE for \texttt{TTDApac} is very high, this happens because the \texttt{TTDApac} leads to an uncertainty map that already has a very high value at every pixel and is not in agreement with the lower per-pixel error (between the prediction and the groundtruth) as evident from Figure 4 and 5 in the main manuscript. As the input sample becomes more noisy (i.e., increasing $\kappa$), the quality of the predictions degrade sharply leading to higher error values, whereas the uncertainty values obtained using \texttt{TTDApac} do not change much. This leads to better agreement between the high error and the uncertainty values causing UCE to decrease with increasing $\kappa$. Moreover, this also indicates that \texttt{TTDApac} does not provide well-calibrated uncertainty estimates.
To shed more light on this phenomena, we also study the UCE trend for two more baseline tasks, where the uncertainty maps are set to constants (0.015 and 0.95). 
For a low-value constant (0.015) uncertainty map, as we degrade the input samples, the quality of output prediction also deteriorates and the disagreement between per-pixel uncertainty and error increases, leading to higher UCE values, therefore there is an increasing UCE trend.
For a high-value constant (0.95) uncertainty map, as we degrade the input samples, the quality of output prediction also deteriorates leading to higher error values which are also closer to higher values of per-pixel uncertainty, leading to more agreement between uncertainty and error and lower UCE values, therefore there is a decreasing UCE trend. Note that this does not indicate well-calibrated uncertainty estimates.
This phenomena is observed for both the super-resolution (Figure~\ref{fig:id}-Left) and deblurring (Figure~\ref{fig:id}-Right).

\subsection{Necessity of the Identity Mapping Term.} 
We ablate our \texttt{BayesCap} by removing the identity mapping loss, i.e., by using the following loss to train \texttt{BayesCap}
\begin{gather*}
\phi^*
= 
\underset{\phi}{\text{argmin}}
\sum_{i=1}^{i=N}
\underbrace{
\left(
\frac{|\tilde{\mathbf{y}}_{i}-\mathbf{y}_{i}|}{\tilde{\alpha}_{i}} \right)^{\tilde{\beta}_{i}} - 
    \log\frac{\tilde{\beta}_{i}}{\tilde{\alpha}_{i}} + \log\Gamma(\frac{1}{\tilde{\beta}_{i}})
}_{\text{Negative log-likelihood}}
\label{eq:bayescap}
\end{gather*}
and compare it with SRGAN on the BSD100 dataset for image super-resolution. Our results 0.34/0.028 UCE score and 0.23/0.45 C.Coeff (for ablation/\texttt{BayesCap} resp.) 
indicate that the identity mapping is essential to learn the calibrated uncertainties.
\begin{table}[]
\centering
\begin{tabular}{lcc}
\toprule
Model   & UCE & C.Coeff  \\ \midrule
No Idenity Mapping & 0.34  & 0.23  \\
BayesCap & 0.028 & 0.45  \\ \bottomrule
\end{tabular}
\caption{Ablation study for the task of image super-resolution with and without the identity mapping. }
\end{table}

\subsection{Comparison with Generalized Gaussian Scratch Model.} 
\texttt{Scratch} is the model trained from scratch that corresponds to
Kendall et al.~\cite{kendall2017uncertainties}. For completeness, we also perform experiments with \texttt{Scratch} model modified to predict the parameters of the Generalized Gaussian distribution, where the fidelity loss term is given by the following equation, 
\begin{gather*}
\mathcal{L}_{\text{fidelity}}
= 
\sum_{i=1}^{i=N} 
\left(
\frac{|\hat{\mathbf{y}}_{i}-\mathbf{y}_{i}|}{\hat{\alpha}_{i}} \right)^{\hat{\beta}_{i}} - 
    \log\frac{\hat{\beta}_{i}}{\hat{\alpha}_{i}} + \log\Gamma(\frac{1}{\hat{\beta}_{i}}) + \lambda_1*\mathcal{L}_{\text{content}}
\label{eq:bayescap}
\end{gather*}
where, $\mathcal{L}_{\text{content}}$ is the content loss from~\cite{Ledig_2017_CVPR}
and the total loss function used for training the network is given by
\begin{gather*}
    \mathcal{L}_{\text{total}} = \lambda_2 \mathcal{L}_{\text{fidelity}} + \lambda_3 \mathcal{L}_{\text{adversarial}}
\end{gather*}
Where $\mathcal{L}_{\text{adversarial}}$ is given by,
\begin{gather*}
    \mathcal{L}_{\text{adversarial}} = \sum_{j=1}^{j=M} - \log{D_{\theta_D}({G_{\theta_G}(x^j))}}
\end{gather*}
where, $x^j$ represents $j^{th}$ image and $M$ is the total number of images in the dataset.
For super-resolution on the BSD100 dataset,  the following are the results:
\begin{table}[]
\centering
\begin{tabular}{lcc}
\toprule
Model   & PSNR & UCE  \\ \midrule
GGD-Scratch & 24.78  & 0.033  \\
Scratch & 24.39  & 0.057  \\
BayesCap & 25.16 & 0.028  \\ \bottomrule
\end{tabular}
\caption{Additional model for the task of image super-resolution that models output as Generalized Gaussian Distribution (GGD) }
\end{table}

\vspace{-40pt}
\subsection{Effect of Post-Hoc Calibration Methods.} We apply post-hoc calibration (variance scaling) from \cite{laves2020calibration}, that is,
we find the optimal scale ($s^*$) by optimizing,
\begin{gather*}
    s^* = \underset{s}{\text{argmin }} N\log(s) + \frac{1}{s^2} \sum_{i=1}^{N} \frac{|\hat{\mathbf{y}}_{i}-\mathbf{y}_{i}|^2}{\hat{\sigma_i}^2}
\end{gather*}
We then rescale the derived variances using the optimal scale ($s^*$). We 
find that the calibration of other models remains worse compared to our \texttt{BayesCap} as shown in the following for super-resolution task on BSD100: 
\begin{table}[]
\centering
\begin{tabular}{lcc}
\toprule
Model   & UCE & C.Coeff  \\ \midrule
Scratch & 0.036  & 0.41  \\
BayesCap & 0.028 & 0.45  \\ \bottomrule
\end{tabular}
\caption{Variance scaling for post hoc uncertainty calibration for the task of image super-resolution}
\end{table}

\subsection{Additional Metrics for Calibration.} We additionally include the \textit{Expected Calibration Error}
(ECE), Sharpness and \textit{Negative Log Likelihood} (NLL) metrics comparing our \texttt{BayesCap} with \texttt{Scratch} and \texttt{DO} respectively for image super-res. on BSD100  
to measure calibration of the uncertainties~\cite{zhou2020estimating,chung2021beyond}, as shown below:
\begin{table}[]
\centering
\begin{tabular}{lccc}
\toprule
Model  & ECE & Sharpness & NLL \\ \midrule
BayesCap & 0.83 & 1.65 & 0.21 \\
Scratch & 1.26  & 2.55 & 0.35  \\
DO & 4.47 & 8.41 & 0.47  \\ \bottomrule
\end{tabular}
\caption{Additional metrics for uncertainty calibration for the task of image super-resolution}
\end{table}

\section{Application: Out-of-Distribution Analysis}
In the main paper, we provide an application for the derived uncertainties from our method to detect OOD samples in depth estimation task.
We used a model trained with KITTI dataset (i.e., MonoDepth2) and evaluate the model on data from KITTI test set (i.e., in distribution), test set for Indian Driving dataset (i.e., out of distribution), and also on test set of Places365 dataset (i.e., severely out of distribution). We used the following methods to detect OOD samples.

\textbf{Using \textit{pretrained} features for OOD detection.}
We computed the mean features for the KITTI validation set images using the feature extracted from the intermediate layer of pretrained MonoDepth2 model, say $\mathscr{M}$, with intermediate feature represented by $\mathscr{M}_l(\cdot)$ (shown below),
\begin{gather}
    \mathcal{F}_{\text{mean}} = \frac{1}{|\text{KITTI val}|} \sum_{i=1}^{i=|\text{KITTI val}|}\mathscr{M}_{l}(x_i) \text{ } \forall x_i \in \text{KITTI val}
\end{gather}
Then, at inference we extract the same intermediate feature for the image being analysed and compute the L2 distance between the KITTI validation set mean feature and the features of the analysed images.
\begin{gather}
    f_{t} =  \mathscr{M}_{l}(x_t) \text{ } \forall x_t \in \text{Inference set} \\
    d_t = ||f_t - \mathcal{F}_{\text{mean}}||^2 \\
    \text{is } x_t \text{ OOD?} = 
    \begin{cases}
    \text{True},& \text{if } d_t\geq \tau\\
    \text{False},              & d_t < \tau
\end{cases}
\end{gather}

\textbf{Using \textit{autoencoder} features for OOD detection.}
We computed the mean features for the KITTI validation set images using the feature extracted from the bottleneck layer of the autoencoder (say $\mathscr{A}$, with bottleneck feature represented by $\mathscr{A}_b(\cdot)$) trained on top of a pretrained MonoDepth2 model, i.e.,
\begin{gather}
    \mathcal{F}_{\text{mean}} = \frac{1}{|\text{KITTI val}|} \sum_{i=1}^{i=|\text{KITTI val}|}\mathscr{A}_b(\mathscr{M}(x_i)) \text{ } \forall x_i \in \text{KITTI val}
\end{gather}
Then, at inference we extract the same bottleneck feature for the image being analysed and compute the L2 distance between the KITTI validation set mean feature and the features of the analysed images.
\begin{gather}
    f_{t} =  \mathscr{A}_{b}(\mathscr{M}(x_t)) \text{ } \forall x_t \in \text{Inference set} \\
    d_t = ||f_t - \mathcal{F}_{\text{mean}}||^2 \\
    \text{is } x_t \text{ OOD?} = 
    \begin{cases}
    \text{True},& \text{if } d_t\geq \tau\\
    \text{False},              & d_t < \tau
\end{cases}
\end{gather}

\textbf{Using \textit{mean uncertainty} for OOD detection.}
At inference, we computed the mean uncertainty values for the images in the inference set using the \texttt{BayesCap} (say $\mathscr{B}$, with uncertainty map represented by $\mathscr{B}_u$) trained on top of pretrained MonoDepth2 model, and use that to decide if a sample is OOD, i.e.,
\begin{gather}
    u_{t} = \text{mean}(\mathscr{B}_{u}(\mathscr{M}(x_t))) \text{ } \forall x_t \in \text{Inference set} \\
\text{is } x_t \text{ OOD?} = 
    \begin{cases}
    \text{True},& \text{if } u_t\geq \tau\\
    \text{False},              & u_t < \tau
\end{cases}
\end{gather}

\bibliographystyle{splncs04}
\bibliography{egbib}